%% file: root.tex
\documentclass[letterpaper, 10 pt, conference]{ieeeconf}  

\IEEEoverridecommandlockouts                              

\usepackage{url}      
\usepackage{graphicx}
\usepackage{amsmath}
\usepackage{amssymb}
\usepackage{algorithm}
\usepackage{algpseudocode}
\usepackage{color,soul}
\usepackage{siunitx}
\usepackage{definitions}
\usepackage[absolute,overlay]{textpos}

\usepackage[colorinlistoftodos, english, disable]{todonotes} 

\usepackage[utf8]{inputenc}

\usepackage[font=footnotesize]{caption}
\usepackage[font=footnotesize]{subcaption}

\usepackage{hyperref}

\title{\LARGE \bf
Decentralized Cooperative Planning for Automated Vehicles with Hierarchical Monte Carlo Tree Search
}

\author{Karl Kurzer$^{*1}$, Chenyang Zhou$^{*2}$ and J. Marius Z\"ollner$^{1,2}$ 
	\thanks{
		$^{1}$Karlsruhe Institute of Technology, Kaiserstr. 12, 76131 Karlsruhe, Germany
		{\tt\small \{kurzer\}@kit.edu}
		$^{2}$FZI Research Center for Information Technology, Haid-und-Neu-Str. 10-14, 76131 
		Karlsruhe, Germany {\tt\small \{czhou, zoellner\}@fzi.de} | \bfseries{*These authors 
		contributed equally.}
	}
}

\graphicspath{{03_graphics/}}
		
\begin{document}
\begin{textblock*}{\textwidth}(19mm,10mm)
	\footnotesize
	\noindent\textcopyright2018~IEEE. Personal use of this material is permitted. Permission from IEEE must be obtained for all other uses, in any current or future media, including reprinting/republishing this material for advertising or promotional purposes, creating new collective works, for resale or redistribution to servers or lists, or reuse of any copyrighted component of this work in other works.\\
	\textit{2018 IEEE Intelligent Vehicles Symposium (IV)}
\end{textblock*}

\maketitle
\thispagestyle{empty}
\pagestyle{empty}

\begin{abstract}
Today's automated vehicles lack the ability to cooperate implicitly with others.
This work presents a Monte Carlo Tree Search (MCTS) based approach for decentralized cooperative planning
using macro-actions for automated vehicles in heterogeneous environments. 
Based on cooperative modeling of other agents and Decoupled-UCT (a variant of MCTS), the algorithm evaluates the
state-action-values of each agent in a cooperative and decentralized manner,
explicitly modeling the interdependence of actions between traffic participants.
Macro-actions allow for temporal extension over multiple time steps and increase the effective search depth requiring
fewer iterations to plan over longer horizons. 
Without predefined policies for macro-actions, the algorithm simultaneously learns policies over and within macro-actions.
The proposed method is evaluated under several conflict scenarios, showing that the algorithm
can achieve effective cooperative planning with learned macro-actions in heterogeneous environments.

\end{abstract}

\section{Introduction}
While the quality of automated driving is progressing at a staggering pace, today's automated vehicles lack a key
ingredient that heavily separates them from their human counterparts --- implicit cooperation. In contrast to the
traditional egoistic maneuver planning methods for automated vehicles, human drivers take other drivers' subtle actions
into consideration enabling them to make cooperative decisions even without explicit communication.

Thus, in recent years a variety of cooperative planning approaches for vehicles have been proposed that take the
interdependence of ones own action and the actions of others into account \cite{Bahram2016,Elvik2014}.
This problem can be treated as a multi-agent Markov Decision Process (MDP).
When ignoring any execution uncertainty, transitions from the current to the next state are fully 
deterministic given the joint actions of all players.
Algorithms designed for single-agent systems often suffer from the \emph{curse of dimensionality} 
when applying them to multi-agent systems, meaning that the number of possible outcomes increases 
exponentially as the number of agents grows, which is even more severe when planning for longer 
time horizons.

\begin{figure}
	\centering
	\def\svgwidth{\columnwidth}
	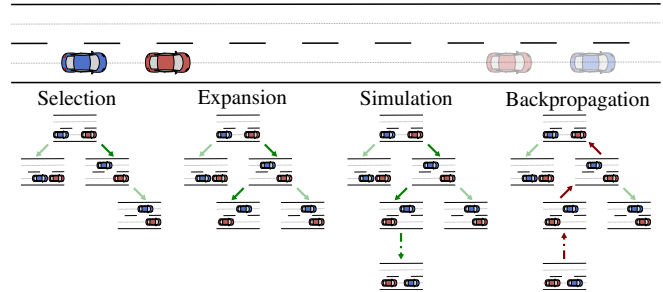
	\caption{Phases of Monte Carlo Tree Search for an overtaking maneuver;
	the selection phase descends the tree by
	selecting promising children until a node is encountered that has not yet been fully expanded.
	Upon expansion random actions are simulated until the planning horizon is reached.
	The result is backpropagated through all nodes along the chosen path.
	Eventually the algorithm converges to an optimal action sequence.}
	\label{Fig:MCTS}
\end{figure}

Monte Carlo Tree Search (MCTS), a reinforcement learning method \cite{Vodopivec2017}, has shown 
promising results on multiple occasions facing problems of this kind. 
The most popular example is the software AlphaGo, reaching super-human performance in the game of Go (\cite{Silver2016,Silver2017}).
MCTS repeatedly samples a model to improve value estimates of actions at a given state 
backpropagating the results from the simulation.
The results guide the selection and expansion phases towards more promising areas of the search space.
An example for the domain of automated driving is given in Fig.~\ref{Fig:MCTS}.
A thorough overview of MCTS and its extensions is presented in \cite{Browne2012}.
Since the performance of MCTS is dominated by its effective search depth \cite{Kearns1999},
and multi-agent problems have an inherently large branching factor, temporal abstraction is used in 
this work by extending actions over several time steps, hereafter macro-actions (MAs).
MAs address the curse of dimensionality by generally reducing the problem complexity, leading to quicker convergence \cite{barto2003recent}.

\HMCTS\ generates \textbf{de}centralized \textbf{co}operative \textbf{h}ierachical plans and applies it to the domain of automated driving.
First, we address the problem of decentralized simultaneous decision making 
with Decoupled-UCT (a variant of MCTS \cite{Tak2014}), removing dependencies on the decisions 
of others.
Additionally, to achieve longer planning horizons we integrate temporally extended macro-actions (MAs) in Decoupled-UCT.
These MAs are designed in a flexible way that requires only initial and terminal conditions to be defined, 
allowing the algorithm to simultaneously learn which MA to choose and how to execute it.
Last we evaluate the capabilities of \HMCTS\ in simulation, showing that it can
achieve effective cooperative planning with learned macro-actions in heterogeneous environments.
In addition the comparison with flat MCTS indicates that our algorithm can generate feasible plans in complex traffic
scenarios with fewer iterations but higher quality.

The remainder of this paper is structured as follows:
Section II gives a brief overview of research on cooperative automated driving, as well as hierarchical reinforcement learning.
The problem is formally defined and its terminology is introduced in section III.
Adaptations to flat MCTS are presented in section IV and the resulting \HMCTS\ algorithm is described in detail in
Section V. 
Lastly, \HMCTS\ is evaluated in a variety of scenarios.

\section{Related Work}

\subsection{Cooperative Driving}
Instead of assuming traffic participants follow merely their own agenda, cooperative planning 
considers others' anticipations and reactions to the ego vehicle's behavior and chooses actions 
that are cooperative, increasing the total utility (\cite{Axelrod1981,During2014}). 
The first successful demonstration of cooperative automated vehicles emerged from the California 
PATH program \cite{swaroop1996string} in the 1990s, where the notion of string stability was 
introduced to maintain the stability of a group of automated vehicles.
Later, other projects focusing on the potential of cooperative perception and motion planning as 
well as the required software and hardware structures were conducted \cite{stiller2007cooperative}.

A definition of cooperative driving behavior, its necessary preconditions and an algorithm to 
generate cooperative plans
is presented in (\cite{During2014, Pascheka2015a}).
Using a utility focused approach cooperative behavior is achieved by increasing the overall 
utility consisting of the sum of each agent's own utility. 
Assuming that utilities of all agents can be perfectly estimated, the presented algorithm finds the 
combination of actions with maximum utility through an exhaustive search.
Actions of agents are represented by quintic polynomials optimized for safety, energy, time, and comfort \cite{Takahashi1989}.

The potential of MCTS for cooperative driving is first presented in \cite{Lenz2016}.
Based on Information-Set MCTS presented in (\cite{CowlingPeterI.2012, Soemers2014})
they ensure decoupled decision making and conduct decentralized planning.
Similar to (\cite{During2014, Pascheka2015a}) they define a set of high-level actions resembled by 
primitive actions with duration of one second.
The algorithm is demonstrated in three different merge scenarios, with up to three vehicles
directly interacting with the ego vehicle, while other vehicles are merely guided by an Intelligent Driver Model \cite{Lenz2016}.
Additionally, the number of lane changes within the planning horizon is restricted to one.

\subsection{Hierarchical Reinforcement Learning}
To address the problem of combinatorial explosion when planning for longer horizons, temporally extendable actions have been long studied in the domain of reinforcement learning. 
Sutton et al. \cite{Sutton1999a} provided a comprehensive framework incorporating temporal abstraction into reinforcement learning.
While they pointed out that there will be a loss of optimality due to the fixed internal structure of the options,
they presented intra-option learning methods (\cite{Sutton1999a, sutton1998intra}) to achieve more flexible options. 

Another framework for hierarchical reinforcement learning was presented in \cite{dietterich2000}, where the entire task
is decomposed hierarchically and then solved by dealing with multiple smaller tasks.
Policies of all elements in the hierarchy can be learned simultaneously. 
The state-action-value function can be recursively decomposed into combinations of the state-action-value of primitive actions.
This algorithm is proved to converge to the recursive optimality.

The combination of MCTS and macro-actions can be done in two ways. Either pre-defined/offline learned MAs represent
the action space of the MCTS or online MA-learning is conducted within the search process. 
A trivial definition of MAs is the repetition of actions, which delivered good results, but cannot be generalized (\cite{Powley2012,Perez2014}).
\cite{de2016monte} creates more complex MAs including domain knowledge and additional algorithms for the execution of MAs.
MAs can be also learned offline by a DQN \cite{Paxton2017a}.
The resulting MAs are more flexible to an extent than the traditional pre-defined MAs, but are still limited. 
Additional Monte Carlo based planning methods using MAs are presented in (\cite{Vien2015,Bai2016a}),
which adopt the MAX-Q framework and propose a hierarchical MCTS algorithm, where each MA is
learned by a nested MCTS in the larger search tree.

\section{Problem Statement}
We formulate the problem of cooperative planning with MAs as a decentralized Semi-Markov Decision
Process (Dec-SMDP).
At each time step, all agents choose an action simultaneously without knowledge of future actions of others,
receive an immediate reward and transfer the system to the consecutive state.
The reward and the state transition is dependent on all agents' actions.

Formally, a Dec-SMDP is described by a tuple
$\langle \agentspace, \statespace,  \actionspace, \transitionmodel, \reward, \discountfactor\rangle$, where
\begin{itemize}
	\item $\agentspace$ is the finite set of \emph{agents} indexed by $i \in {1, 2, \dots n}$. 
	\item $\statespacei$ is the finite \emph{state space} of an agent, $\statespace = \times \statespacei$ represents
	the joint state space of $\agentspace$. 
	\item $\actionspacei$ is the finite  \emph{action space} of an agent, $\actionspace= \times \actionspacei$
	represents the joint action space of $\agentspace$. 
	\item $\transitionmodel: \statespace \times \actionspace \times \statespace \to [0,1]$ is the
	\emph{transition function} $P(s' | s, \actionj)$ 
	which specifies the probability of the transition from state $s$ to state $s'$
	under the joint action $\actionj$ defined by each agent's choice. 
	\item $\reward: \statespace \times \actionspace \times \statespace \to \mathbb{R}$ is the reward function with
	$r(s, s',\actionj)$ representing the reward after the joint action $\actionj$ is executed. 
	\item $\discountfactor \in [0,1]$ is a \emph{discount factor} which controls the influence of future rewards on the
	current state. 
\end{itemize}

We use superscript $^i$ to denote that a parameter is related to agent $i$.
The solution to the Dec-SMDP is the joint policy $\Pi = \langle \pi^1, \dots, \pi^n \rangle$, where $\pi^i$ denotes the
individual policy for a single agent, i.e., a mapping from the state to the probabilities of each available action,
$\pi^i: \statespace^i\times \actionspace^i \to [0,1]$. 
Each agent tries to maximize its expected cumulative reward starting from its current state:
$G = \sum {\gamma^t r(s ,s', a)}$,
where $t$ is the time and $G$ is the return, representing the cumulated discounted reward. 
$V(s)$ is called the state-value function, given by $V^\pi(s) = E[G|s,\pi]$.
Similarly, the state-action-value function $Q(s,a)$ is defined as $Q^\pi(s,a) = E[G|s,a]$,
representing the expected return when choosing action $\action$ in state $\state$.

The optimal policy starting at state $s$ is defined as $\pi^* = \argmax_{\pi} V^\pi(s)$. 
The state-value function is optimal under the optimal policy: $\max V = V^{\pi^*}$, the same is 
true for the
state-action-value function: $\max Q=Q^{\pi^*}$.
The optimal policy can be found by maximizing over $Q^*(s,a)$:
\begin{equation}\label{Eq:OptPolicy}
	\pi^{*}(a|s) = \left\{ \begin{array}{rcl}
	1 & {\mbox{if}} \ {a = \argmax_{a \in {\cal A}} Q^{*}(s,a)}  \\
	0 & \mbox{otherwise}
		   \end{array}\right.
\end{equation}
It should be mentioned that $Q^{*}$ is stored in a table, i.e., each discrete action is assigned with its state-action-value $Q(s,a)$. Once $Q^*$ has been determined, the optimal policies can easily be derived.
Thus the goal is transformed to learning the optimal state-action-value function $Q^*(s,a)$ for each state-action
combination (macro/primitive).

Compared to the single level of policies in an MDP, there exists a hierarchy of policies in an SMDP where each MA $\omega$
has its own policy $\pi_{\omega}$
and a policy $\pi_\mu$ which decides how to choose the next MA $\omega'$ when the previous one terminates.
The Bellman equations of the SMDP can be written as:
\begin{equation}\label{Eq:VQSMDP}
		\begin{split}
	Q^{\pi_{\mu}}(s,\omega) &= \sum^{\tau}_{k=1} {\gamma^{k-1} r_{t+k}}\\
	&+\sum_{s',\tau}{\gamma^{\tau} 
	p(s',\tau|s,\omega)\sum_{\omega'}{\pi_{\mu}(\omega'|s')Q^{\pi_{\mu}}(s',\omega')}}
	\end{split}
\end{equation}
The first part is the cumulated reward for this MA $\omega$ during its execution for $\tau$ steps.
The latter part is the completion term ${C}$ which can be further decomposed until primitive actions are encountered \cite{dietterich2000}.
Thus the state-action-value of choosing MA $\omega$, $Q^{\mu}(s,\omega)$ can be viewed as the cumulative discounted reward by following
the policy of the chosen MA $\pi_\omega$ and then the policy $\pi_\mu$ which chooses this MA until $\pi_\mu$ ends.

\section{Approach}
\subsection{Hierarchical Action Graph}
This section presents our design of macro-actions with a hierarchical graph for the cooperative driving domain,
based on the \emph{Option} \cite{Sutton1999a} and \emph{MAXQ} \cite{dietterich2000} frameworks.

While MAs reduce the complexity and thus the search space, the following are key challenges that must be addressed
when implementing MAs.

\subsubsection{Asynchronous decision making}
In a multi-agent system with variable duration of MAs, MAs end asynchronously. 
Different strategies are presented in \cite{MahadevanSummary}.
\begin{itemize}
	\item $t_{all}$ keeps some agents idle to wait for others finishing their MAs
	\item $t_{any}$ simply interrupts all MAs when the first agent finishes its MA
	\item $t_{continue}$ allows asynchronous selection of MAs, which means that each agent independently decides its
	next macro-action once it terminates its current MA.
\end{itemize}

Clearly, the first two schemes $t_{all}$ and $t_{any}$ force a synchronization of the decision epochs and can
only be realized in a centralized way, while $t_{continue}$ allows decentralized asynchronous decision making.
\subsubsection{Flexible design of MAs}
Naive pre-defined MAs are even more harmful than only planning with primitive actions \cite{Sutton1999a}.
To mitigate the risk, the policy inside a MA should be learned online, allowing for a flexible 
adaptation to a given situation.

\subsubsection{Cooperation Level}
Learning of MAs can be distracted by lower level actions of other agents in a multi-agent system \cite{Ghavamzadeh2006}.
Consequently, approaches such as localized macro-actions that do not consider cooperation at the 
level of primitive actions are developed in (\cite{Liu2017, Amato2014a, 
omidshafiei2015decentralized}).
However, they are unsuitable for cooperative automated driving,
where consideration of primitive actions of all agents is required, e.g., for collision checking.

The \emph{Option} framework \cite{Sutton1999a} generalizes the primitive actions into temporally extended MAs
with three components $\langle I, \pi, \beta\rangle$, where the MA is referred to as the option. 
$I$ is the \emph{initiation set} which specifies if this MA is available at the current state. 
$\pi: \statespace \times \actionspace \to [0,1]$ is the above mentioned \emph{policy} for this MA. 
$\beta: \statespace \to [0,1]$ is the \emph{termination probability} that the MA terminates at the current state. 
Note that for primitive actions these three components are defined as $\pi(s,a) = 1$, $\beta (s)=1$ and $I = S$. 

The \emph{MAXQ} framework \cite{dietterich2000} formulates the whole scenario as a root task and decomposes the root task into sub-tasks. 
To solve the root task, sub-tasks are sequentially chosen according to the root policy $\pi_{\mu}$ and the sub-tasks
are solved according to their own policies $\pi_{\omega}$. 
The sub-tasks can be further decomposed into sub-sub-tasks until a primitive task (action) is encountered,
the policies are decomposed accordingly. 

\begin{figure}
	\centering
	\def\svgwidth{\columnwidth}
	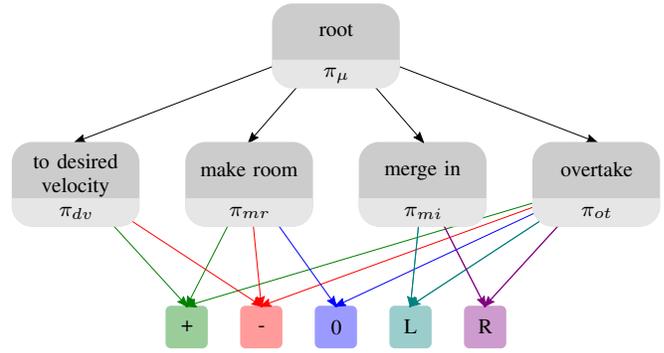
	\caption{The hierarchical action graph originates from an abstract root macro-action $\mu$ with a policy $\pi_\mu$
	that selects other abstract macro-actions $\omega$.
	These MAs have different action sets that are defined by primitive actions, that the agent can execute.}
	\label{Fig:ActionGraph}
\end{figure}

\begin{figure}
	\centering
	\def\svgwidth{\columnwidth}
	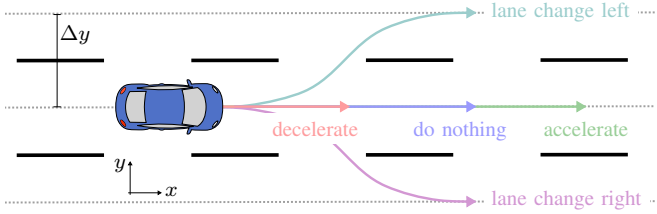
	\caption{Trajectories of the five primitive actions}
	\label{Fig:TrajFig}
\end{figure}

We adopt the hierarchical action graph from the \emph{MAXQ} framework and present our hierarchical action graph in Fig.~\ref{Fig:ActionGraph}. 
Considering the listed conflict scenarios that require cooperative driving in \cite{Ulbrich2015},
we propose four MAs: \emph{overtake}, \emph{merge in}, \emph{make room}, \emph{to desired velocity}.
Each MA has four components $\langle I, \pi, \actionspace_{\omega}, \beta\rangle$, where 
$\actionspace_{\omega}$ is the set of available actions at the immediate lower level. 
The primitive actions are defined as \emph{acceleration}, \emph{deceleration}, \emph{do-nothing}, \emph{lane change left} and \emph{lane change right}.
They are represented by quintic polynomials describing the changes in longitudinal velocity $\Delta 
\dot x$ and lateral position $\Delta y$ (\cite{During2014, Takahashi1989}), depicted in 
Fig.~\ref{Fig:TrajFig}.
The position of a vehicle refers to the lateral and longitudinal position of the midpoint of the 
rear axle in the world coordinate.
Each MA has a subset of these primitive actions and is referred to as the parent action of the 
primitive actions.
The solution to the general driving task is generalized as the \emph{root} MA $\mu$ that entails all lower MAs.
The initiation set $I$ (or initial condition) and termination probability $\beta$ (or termination condition) of all
macro-actions are defined in Table~\ref{Table:InitialTerminalConditions}.

\begin{table}
	\caption{Initial and Terminal Conditions for Macro-Actions}
	\centering
	\begin{tabular}{||p{1.5cm} p{2.75cm} p{2.8cm}||} 
	 \hline
	 Macro-Action & Initial Condition & Terminal Condition \\ [0.3ex] 
	 \hline\hline
	 overtake & behind slower vehicle and left lane exists & ahead of slower vehicle\\ 
	 \hline
	 merge in & not in desired lane & in desired lane\\
	 \hline
	 make room & always possible & always possible \\
	 \hline
	 to desired velocity & not at desired velocity & at desired velocity\\ [0.5ex] 
	 \hline
	\end{tabular}
	\label{Table:InitialTerminalConditions}
\end{table}

\begin{figure}
	\centering
	\def\svgwidth{\columnwidth}
	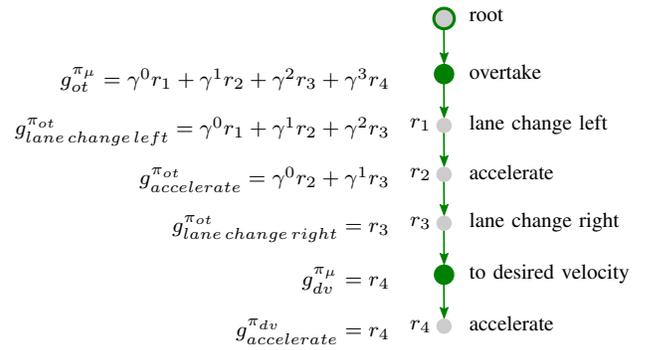
	\caption{Hierarchically bounded return for the example of a single agent; the returns of this iteration are defined
	on the left side.
	Note that the return for choosing MA \emph{overtake} under the root policy $\pi_{\mu}$ includes all rewards from $r_1$ to $r_4$,
	since $\pi_{\mu}$ terminates at $t=4$, while the return for choosing \emph{lane change left} under policy $\pi_{ot}$
	only includes the reward from $r_1$ to $r_3$ because the MA \emph{overtake} terminates at $t=3$.
	This means that the return of an action $a$ is bounded within its parent MA $\omega$.}
	\label{Fig:BoundedReturn}
\end{figure}

As defined by \eqref{Eq:VQSMDP}, the value of choosing action $a$ at state $s$ according to policy 
$\pi$ is the
cumulative discounted reward starting from the current state until $\pi$ ends.
An example for one iteration in the single agent domain with \emph{hierarchically bounded return} is depicted
in Fig.~\ref{Fig:BoundedReturn}.

\subsection{Decision Making without Communication}
Since agents are not communicating, decentralized planning needs to be conducted, where each agent can only influence
its own action, rather than the joint action of all agents. 
The state-action-value estimation of agent i, $Q(s,a^i)$, cannot distinguish among all joint actions $\boldsymbol{a}$ containing this agent's action.
Based on the idea from simultaneous games \cite{Tak2014} and distributed reinforcement learning
area \cite{Lauer2000},
we conduct the marginalization over these joint actions, see \eqref{Eq:Dec}:
 \begin{equation}
 	\label{Eq:Dec}
		Q(s,a^{i}) = \frac{1}{N(s,a^{i})}\sum_{\boldsymbol{a}}{1(a^{i}) 
		N(s,\boldsymbol{a})Q(s,\boldsymbol{a})}
		,
 \end{equation}
where $N(s,a^i)$ and $N(s,\boldsymbol{a})$ represent visit count of agent $i$'s action $a$ and the joint action $\boldsymbol{a}$ respectively.

\subsection{Cooperative Reward Function} \label{sec:RewardFn}
As opposed to classical multi-agent systems with an explicit common goal and an immediate reward for the joint action,
the common goal for cooperative driving is rather implicitly stated --- solving a scenario with conflicting interests
maximizing the overall reward, given each vehicle's safety, efficiency and comfort preferences.
Similar to (\cite{During2014, Lenz2016}) \HMCTS\ assumes identical reward functions for all agents.
For each agent $i$ a cooperative reward \cooperativeReward\ is calculated, which is the sum of its own reward $r_{i}$
according to \eqref{Eq:Reward} as well as the rewards of all other agents based on 
\eqref{Eq:CooperativeReward}.

\begin{equation}\label{Eq:Reward}
	\begin{aligned}
	r^i &= r^i_{\phi} + r^i_{action} \\
		&= r^i_{\phi} + r^i_{safety} + r^i_{efficiency} + r^i_{comfort}\\
	\end{aligned}
\end{equation}
\begin{equation}\label{Eq:CooperativeReward}
	\cooperativeReward = r^i + \lambda \sum_{j = 0, j \neq i}^{n}r^{j}
\end{equation}

$r^i_{\phi}$ is the shaping term described in the next section. $\lambda^i \in [0,1]$ is a cooperation
factor that determines the agent's willingness to cooperate with other agents
(from $\lambda^i = 0$ \emph{egoistic}, to $\lambda^i = 1$ \emph{fully cooperative}).
With the goal to generate cooperative maneuver decisions $\lambda^i$ should be larger than 0.

\subsection{Reward Shaping} \label{sec:rewardShaping}
Potential based reward shaping is used to accelerate the convergence of the learning process, while being optimality
invariant \cite{ng1999policy}.
A potential function $\phi (s)$ is defined to determine the potential of each state. 
The closer the current state to the desired state is, the higher the potential will be.
Our work describes a desire that an agent strives to fulfill as a certain velocity and lane index 
to be reached.
In an MDP, the potential based reward for a transition from state s to s' by action $a$ is written as:

\begin{equation}\label{Eq:PotentialBasedReward}
	r_{\phi}(s,s',a) = \gamma \phi (s') - \phi (s)
\end{equation}
 Thus, the ego reward function for each agent can be written as:
 \begin{equation}\label{Eq:PotentialBasedReward_i}
	r^i = r^i_{action} +  \gamma \phi (s') - \phi (s)
 \end{equation}

The potential shaping term can be generalized in the SMDP with an additional parameter $\tau$ denoting the duration of the MA,
defined as:
\begin{equation}\label{Eq:ShapingOptionDef}
	r_{\phi}(s,s_{t+\tau}, \omega) = \gamma^{\tau} \phi (s_{t+\tau}) - \phi (s_t)
\end{equation}
It can be proved that \eqref{Eq:ShapingOptionDef} is equivalent to the discounted sum of the shaped 
terms for each
primitive action within its MA $\omega$:
\begin{equation}\label{Eq:ShapingOptionPrac}
	\gamma^{\tau} \phi (s_{t+\tau}) - \phi (s_t) = \sum^{\tau}_{k=1} {\gamma^{k-1} r_{\phi}(s,s_{t+\tau}, \omega)^{a_{t+k}}}
\end{equation}

\section{Algorithm}
\begin{algorithm}
		\caption{\HMCTS}
		\label{Alg:HMCTS}
\begin{algorithmic}
	\Function{Planning}{$\agentspace, \actionspace, \statej$}
	\State $\actionj \gets \emptyset$
		\While{driving}
			\State new root node $n_\mu \gets n\langle \actionj, \agentspace, \actionspace, \statej \rangle$
			\State $a\gets $ \HMCTS($n_\mu$)
			\State $\statej\gets$ ExecuteAction$(a)$
		\EndWhile
	\State \textbf{return}
\EndFunction
\\
\Function{\HMCTS}{$n_\mu$}
	\While{computational budget not reached}
		\State $\langle \nodel, \mathcal{R}\rangle\gets $TreePolicy$(n_\mu)$
		\State $\mathcal{R} \gets$ SimulationPolicy$(\nodel, \mathcal{R})$
		\State BackpropagationPolicy$(\nodel, \mathcal{R})$
	\EndWhile
	\State \textbf{return} $a \gets $FinalSelection$(n_\mu)$
\EndFunction
\\
\Function{TreePolicy}{$n$}
\label{Fn:TreePolicy}
	\Repeat
		\For{$i = 1 $ to $|\agentspace|$}
			\State $a^i \gets $ UCTAction($n, i$)
			\State $\actionj \gets [\actionj, a^i]$
		\EndFor
		\State $n \gets $SelectNode$(n, \actionj)$
		\State $\mathcal{R}\gets $CollectReward($n, \actionj$)
	\Until{$n == \emptyset$ }
		\State $\langle \nodel, \mathcal{R}\rangle \gets$ ExpandNode($n, \actionj$)
	\State \textbf{return} $\nodel, \mathcal{R}$
\EndFunction
\\
\Function{SimulationPolicy}{$n$}
	\For{$i = 1 $ to $|\agentspace|$}
		\State action $a^i\gets$ RandomSelection$(n, i)$ 
		\State $\actionj \gets [\actionj, a^i]$
	\EndFor
	\State $\mathcal{R}\gets $CollectReward($n, \actionj$)
	\State \textbf{return} $\mathcal{R}$
\EndFunction

\\
\Function{BackpropagationPolicy}{$n, \mathcal{R}$}
	\While{$n \neq n_\mu$}
		\State $N(n)\gets N(n)+1$
		\For{$i = 1 $ to $|\agentspace|$}
			\State  $N(a^i)\gets N(a^i)+1$
			\State $G^i \gets \sum^{a^i_{p}} {\gamma^t r^i}$
			\State  $Q(a^i) \gets Q(a^i) + \frac{G^i - Q(a^i)}{N(a^i)}$
		\EndFor
		\State $n \gets$ $n_{parent}$
	\EndWhile
\EndFunction


\end{algorithmic}
\end{algorithm}
\todo[inline, color = green]{preserved the collective reward with short explaination}
We call our algorithm \HMCTS, its most important functions are outlined in Algorithm~\ref{Alg:HMCTS}. 
It preserves the classical four steps of MCTS: selection, expansion, simulation and backpropagation.
The function \textsc{TreePolicy} contains the selection and expansion steps. 
Like traditional MCTS the algorithm builds a search tree of possible future states,
starting from the root node $\mu$ representing the initial state.

\subsubsection{Tree Policy}
UCT with a single agent expands nodes until all available actions have been tried and then continues to grow the tree deeper.
In the decentralized multi-agent system, the agent cannot distinguish between the joint actions. 
As a result, the tree can grow deeper once each agent has explored all of its available actions once.
As \cite{schaeffer2009comparing} suggests that the deterministic UCT in multi-player games does not necessarily converge to a Nash equilibrium,
$\epsilon$-Greedy is introduced and each agent selects an action with stochastic UCT as follows:
\begin{equation}\label{Eq:EpsilonGreedy}
	\pi^{\epsilon}(a|s) = \left\{ \begin{array}{rcl}
	1 - \epsilon + \frac{\epsilon}{|{\cal A}|} & {\mbox{if}} \ {a = \argmax_{a \in {\cal A}} UCT(a)}  \\
	\frac{\epsilon}{|{\cal A}|} & \mbox{otherwise}
		   \end{array}\right.
\end{equation}

Since only primitive actions receive immediate rewards and can trigger system transitions,
a joint action $\actionj$ is required to contain only primitive actions (not MAs) to be executed and transfer the system to a consecutive state.
This implies that all agents select according to their hierarchical policies until a primitive action is chosen.
Our approach adopts the $t_{continue}$ termination mechanism to deal with MAs of variable duration, i.e.,
the decision making for the next MA is independent of the others' current MAs and thus asynchronous for all agents.

\subsubsection{Simulation Policy}
As no prior knowledge is used, the simulation policy simply chooses MAs and their respective primitive actions at random.

\subsubsection{Backpropagation Policy}
Basic MCTS usually uses the simulation outcome without any intermediate rewards for actions in the backpropagation step \cite{Browne2012}. 
By contrast, the return, i.e., cumulative discounted reward is used in our approach.
As mentioned before, the return for the current action is bounded within its parent action. 
Both (\cite{Vien2015,Bai2016a}) use the recursive form of MCTS based on the POMCP \cite{NIPS2010_4031} to realizes the
hierarchically bounded return, which is only applicable in the single-agent system and the multi-agent system
with $t_{any}$ or $t_{all}$ termination rule.
To combine the hierarchically bounded return with $t_{continue}$ in a multi-agent system,
the rewards along each iteration are stored in a table
together with the corresponding hierarchical information about the action, indicated by $\mathcal{R}$.
We then conduct a hierarchical boundary check based on this reward sequence, determining with which MA the reward is associated.

\subsubsection{Final Selection and Execution}
When the termination conditions are met, the agent has learned a policy hierarchy and chooses an action according to the
max reward or max visits principle.
It should be mentioned that the selected primitive action belongs to a certain macro-action.
When starting a new search, this information can be incorporated in the new tree or discarded,
which is called \emph{hierarchical} control mode and \emph{polling} control mode respectively.
\cite{Sutton1999a} showed that \emph{polling} which starts a new planning cycle without
any memory about the previous step yields better results because polling allows the premature
termination of macro-actions at each step and is thus more flexible.


\section{Evaluation}
The evaluation is conducted using a simulation.
We use three different scenarios to test if our algorithm can meet the following goals:
\begin{itemize}
	\item Learning of MAs
	\item Converging quicker than flat MCTS
	\item Finding robust solutions when encountering non-cooperative drivers
\end{itemize}

Each scenario is defined by initial variable values indexed with $_0$, and desired values indexed by $_{des}$ denoting the agents desire.
$x$ denotes the position, $v$ the velocity and $l$ the lane index respectively.
A video of the algorithm in execution can be found online \footnote{http://url.fzi.de/DeCoH-MCTS-IV}.

\subsection{Learning of Macro-Actions}
\begin{figure}
	\includegraphics[width=\columnwidth]{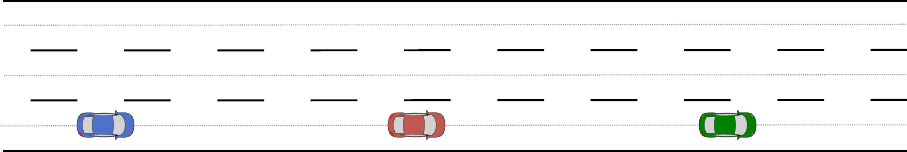}
	\caption{Scenario: Overtake}
	\label{Fig:Overtake3Veh}
\end{figure}

The overtake scenario in Fig.~\ref{Fig:Overtake3Veh} is considered to test the algorithm's ability to simultaneously learn which MA to choose and how to execute it.
Table~\ref{Table:OvertakeScenario} defines the settings for the scenario.
All three vehicles are controlled by their own \HMCTS\ with $\cooperationFactor = 1$. The step length is set to \SI{2}{s}, a total of 2,000 iterations are executed with a maximum planing horizon of 20 steps.
The resulting plan found at step 0 with the given scenario configuration is depicted in Table~\ref{Table:LearningMAOutput}
\begin{table}
	\caption{Configuration for the Overtake Scenario}
	\centering
	\begin{tabular}{||p{0.1cm} p{0.5cm} p{0.8cm} p{0.9cm} p{0.9cm} p{0.9cm} p{1.3cm}||} 
	 \hline
	 ID & color & $x_{0}$ & $v_{0}$ & $l_{0}$ & $v_{des}$ & $l_{des}$ \\
	 \hline
	 0 & blue & \SI{5}{m} & \SI{15}{m/s} & 0 & \SI{25}{m/s} & 0 \\ 
	 1 & red & \SI{25}{m} & \SI{15}{m/s} & 0 & \SI{20}{m/s} & 0 \\ 
	 2 & green & \SI{45}{m} & \SI{15}{m/s} & 0 & \SI{15}{m/s} & 0 \\ 
	 \hline
	\end{tabular}
	\label{Table:OvertakeScenario}
\end{table}

\begin{table}
	\caption{Plan Result at Step 0}
	\centering
	\begin{tabular}{||p{1cm} p{6cm}||} 
		\hline
		Agent & Planned Action Sequences \\ [0.3ex] 
		\hline\hline
		0 & Overtake \\ 
		& \texttt{L L + + 0 0 R R} \\
		\hline
		1 & Overtake \\
		& \texttt{L + + 0 R 0 0 0} \\
		\hline
		2 & Make Room\\ [0.5ex] 
		& \texttt{+ 0 - 0 0 - + 0}\\
		\hline
	\end{tabular}
	\label{Table:LearningMAOutput}
\end{table}

\begin{figure}
	\centering
	\includegraphics[width=0.85\columnwidth]{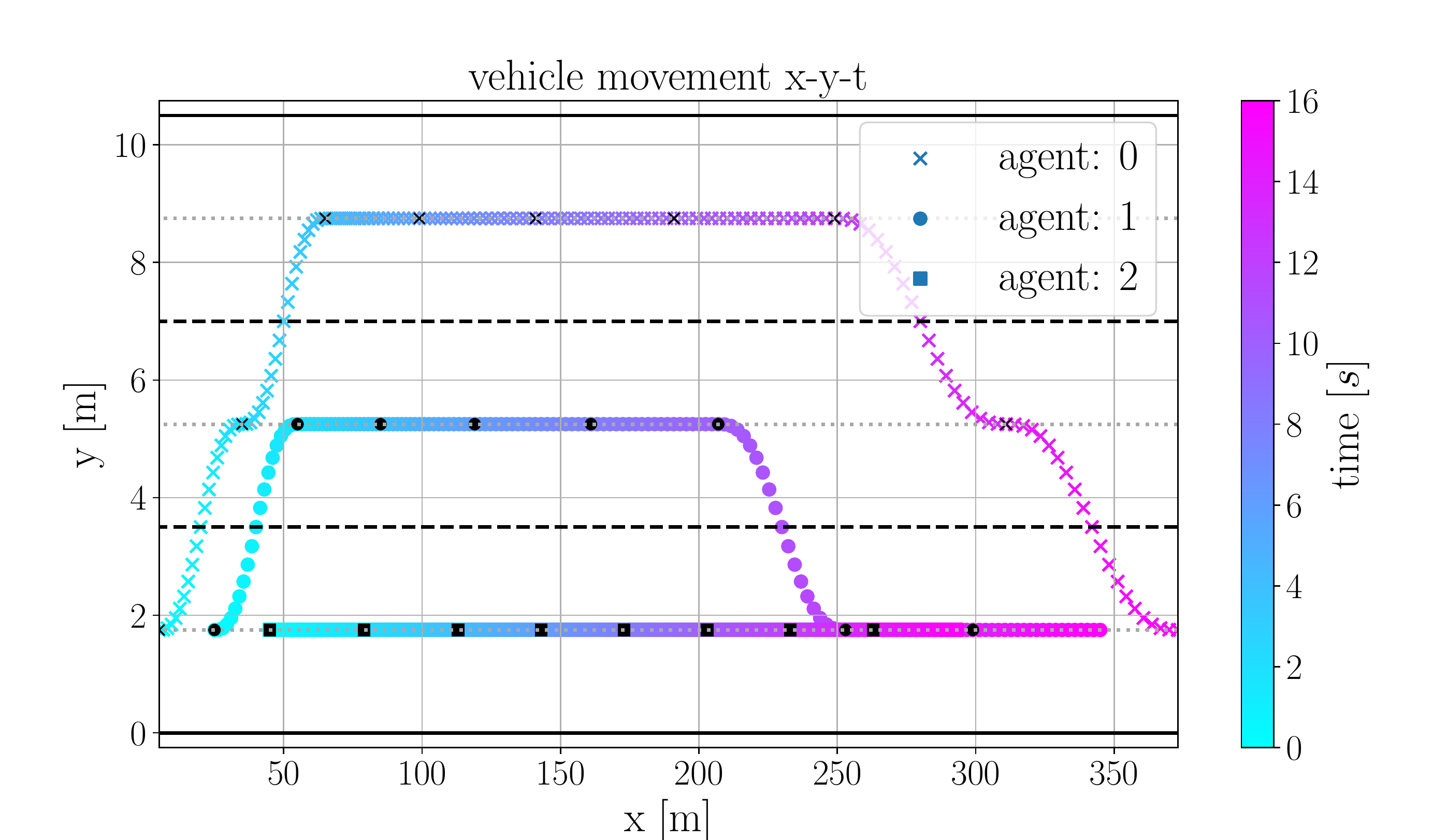}
	\caption{Trajectories for each agent of the overtake scenario;
		The color represents the time according to the color bar on the right side.
		It can be seen that agent 1 changes to the left after driving in front of agent 2,
		while agent 0 stays in lane 2 until it gets in front of both two vehicles
		and then makes two lane changes to its desired lane 0.}\label{Fig:2DOvertake3}
\end{figure}

It can be seen that agent 0 learns the MA \emph{overtake} differently from agent 1,
where agent 0 makes two lane changes to the left, accelerates to get in front of the other two 
vehicles and finally returns back to its desired lane.
Agent 2 shows cooperative behavior by making room for others as it is being tailgated.

In the polling control mode, each agent executes the planned action, i.e., L, L, + respectively,
and then starts a new plan without memorizing the previous result.
Fig.~\ref{Fig:2DOvertake3} shows the 2D trajectories for each agent.

\subsection{Convergence}
\begin{figure}
	\includegraphics[width=\columnwidth]{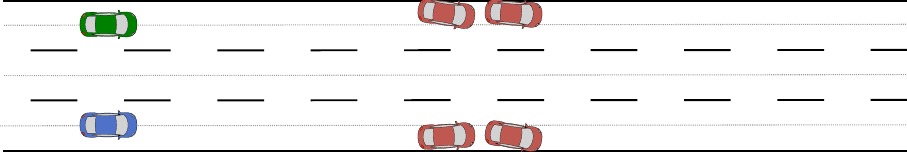}
	\caption{Scenario: Double Merge}\label{Fig:DoubleMerge}
\end{figure}

A test of the convergence speeds between \HMCTS\ and flat MCTS is conducted using the double merge scenario in Fig.~\ref{Fig:DoubleMerge}.
There are two vehicles on a three-lane road with another four parked vehicles blocking the rightmost and leftmost lane.
Both vehicle 0 (blue) and vehicle 1 (red) start with a speed of \SI{25}{m/s} and want to keep their current lane and velocity.

\begin{figure}
	\begin{subfigure}{\columnwidth}
		\centering
		\includegraphics[clip, trim=0 0cm 0 1.75cm, width=0.8\columnwidth]{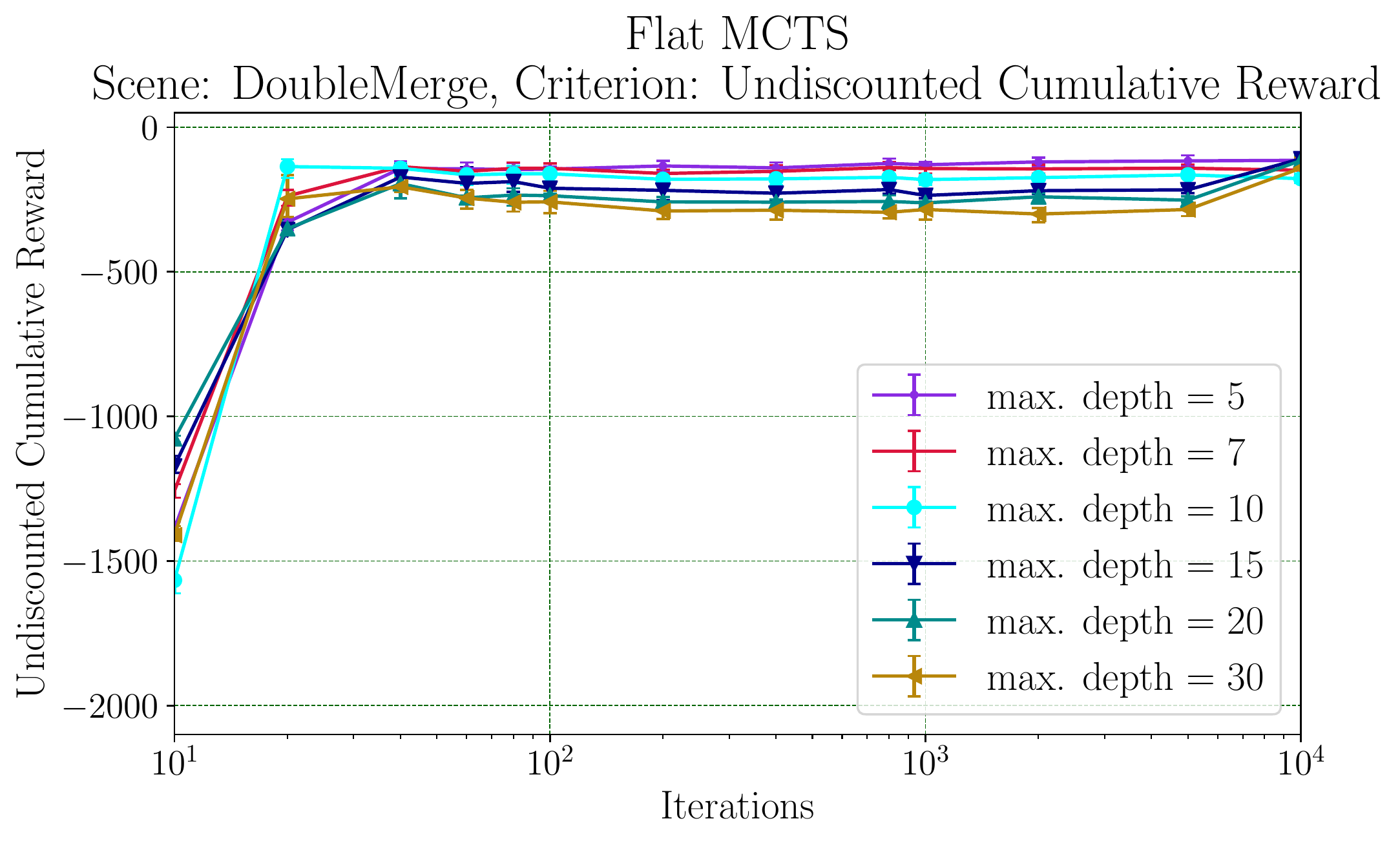}\hfill
		\caption{Undiscounted return of flat MCTS}
		\vspace{0.4cm}
	\end{subfigure}
	\begin{subfigure}{\columnwidth}
		\centering
		\includegraphics[clip, trim=0 0cm 0 1.75cm, 
		width=0.8\columnwidth]{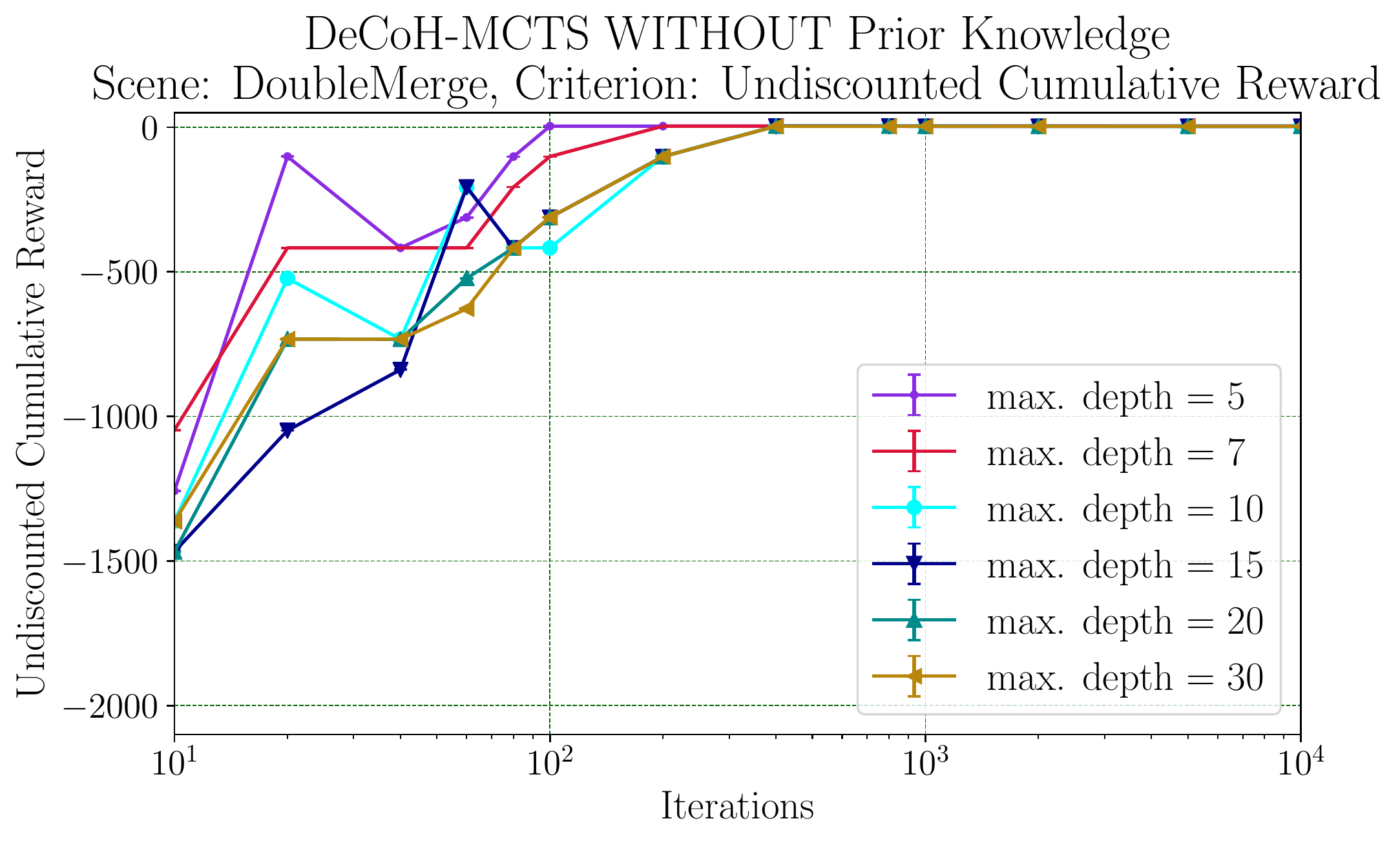}\hfill
		\caption{Undiscounted return of {\HMCTS} \textbf{without} domain knowledge}			
		\vspace{0.4cm}
	\end{subfigure}
	\begin{subfigure}{\columnwidth}
		\centering
		\includegraphics[clip, trim=0 0cm 0 1.75cm, 
		width=0.8\columnwidth]{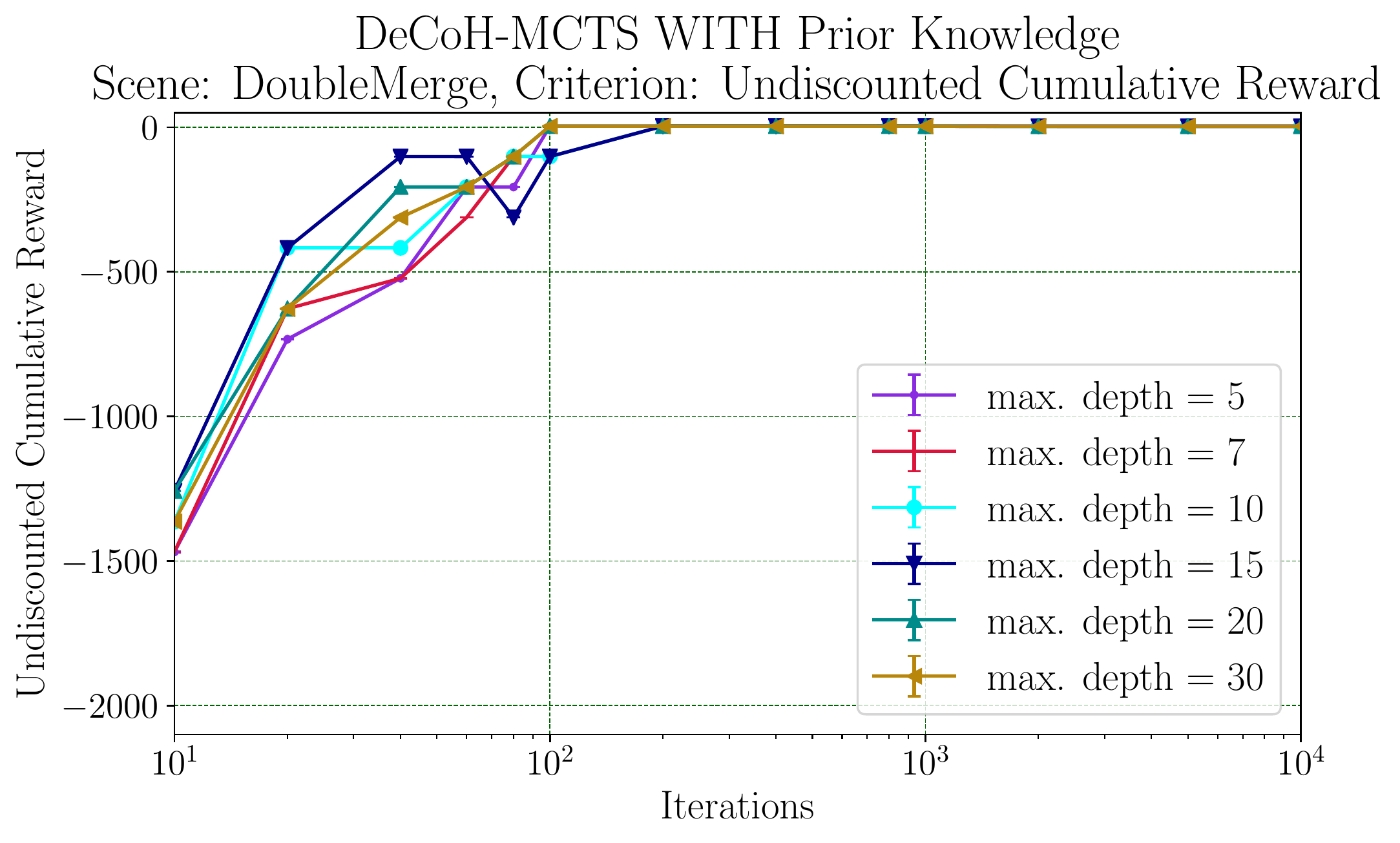}\hfill
		\caption{Undiscounted return of {\HMCTS} \textbf{with} domain knowledge}
		\end{subfigure}
		\caption{Performance comparison of different MCTS versions in the double merge scenario}
		\label{Fig:DoubleMergeConvergence}
\end{figure}

We equip the macro-action overtake in one test with domain knowledge and implement it similar to the
$\epsilon$-greedy policy during the simulation.
The algorithm runs in polling control mode with a step length of 2 seconds.
The undiscounted cumulated rewards \cooperativeReward of vehicle 0 w.r.t. different numbers of iterations and maximal tree depths
are calculated and compared, see Fig.~\ref{Fig:DoubleMergeConvergence}.
Each data point is the mean value of 30 runs with an error bar consisting of upper and lower quartiles.

It can be seen that flat MCTS performs better in terms of undiscounted return for a low number of iterations,
but \HMCTS\ clearly converges to a higher optimum as the number of iterations increases ($>$100).
Considering that the number of iterations lies usually around 1000, the \HMCTS\ performs better than flat MCTS.
It is observed that an increase in the maximal search depth leads to poorer performance for the classical MCTS as opposed to {\HMCTS}. 
The reason is that classical MCTS does not have any domain specific knowledge and conducts the simulation totally random.
Larger maximum search depth means that the random simulations are more likely to end in collisions.
As a result, almost all actions are evaluated negatively.
It becomes difficult for the agent to choose actions that fulfill its desire while not leading to a colliding state,
so that the agent eventually chooses to decelerate to a standstill, being the safest option.

In addition, the comparison between \HMCTS\ with and without domain knowledge shows that the integration of domain knowledge
can accelerate the learning speed while it does not affect the optimality of the solution.

\subsection{Robustness when encountering non-cooperative drivers}
\begin{figure}
	\includegraphics[width=\columnwidth]{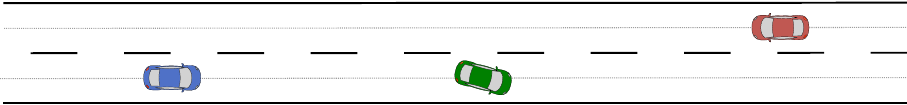}
	\caption{Scenario: Bottleneck, where the green vehicle blocks one lane}\label{Fig:Bottleneck}
\end{figure}

The bottleneck scenario is used to demonstrate the robustness in situations where other agents do not behave as the algorithm assumes (see~\ref{sec:RewardFn}).
As Fig.~\ref{Fig:Bottleneck} and Table~\ref{Table:BottleneckScenario} show, vehicle 0 (blue) approaches from the left and is controlled by {\HMCTS},
vehicle 1 (red) drives from the right at different constant velocities in the range of $v \in [\SI{5}{m/s}, \SI{17}{m/s}]$.
Vehicle 1 keeps its velocity and does not react to vehicle 0 at all.
Vehicle 2 (green) blocks the lane of vehicle 0.

\begin{table}
	\caption{Configuration for the Bottleneck Scenario}
	\centering
	\begin{tabular}{||p{0.1cm} p{0.4cm} p{0.7cm} p{1.0cm} p{0.8cm} p{1.0cm} p{1.3cm}||} 
	 \hline
	 ID & color & $x_{0}$ & $v_{0}$ & $l_{0}$ & $v_{des}$ & $l_{des}$ \\
	 \hline
	 0 & blue & \SI{5}{m} & \SI{10}{m/s} & 0 & \SI{15}{m/s} & 0 \\ 
	 1 & red & \SI{195}{m} & 5-17m/s & 0 & 5-17m/s & 1 \\ 
	 2 & green & \SI{100}{m} & \SI{0}{m/s} & 0 & \SI{0}{m/s} & 0 \\ 
	 \hline
	\end{tabular}
	\label{Table:BottleneckScenario}
\end{table}

\begin{figure}
  \begin{subfigure}{0.49\columnwidth}
  \includegraphics[width=\columnwidth]{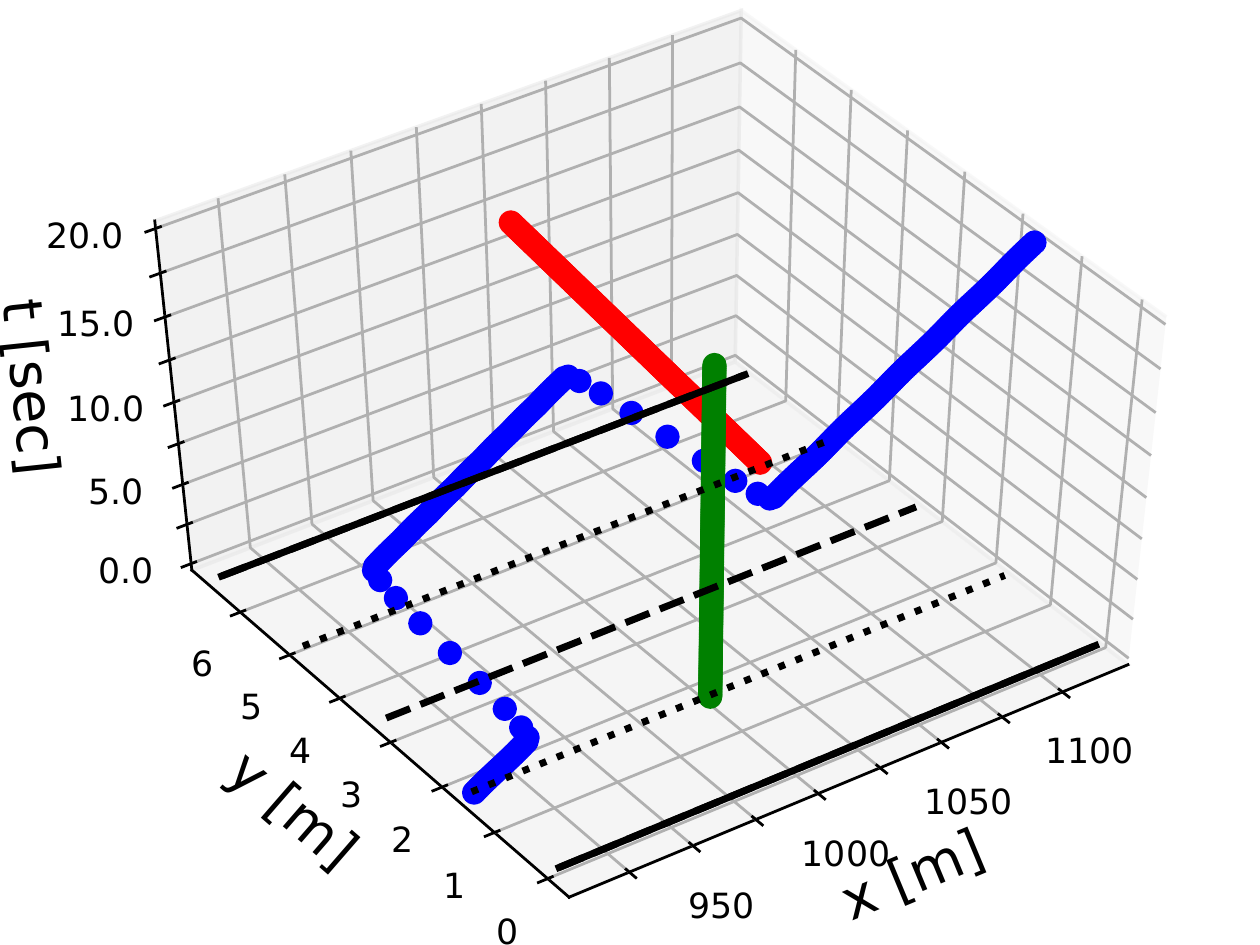}
  \caption{$v_{red} = \SI{5}{m/s}$}
  \end{subfigure}
  \begin{subfigure}{0.49\columnwidth}
  	\includegraphics[width=\columnwidth]{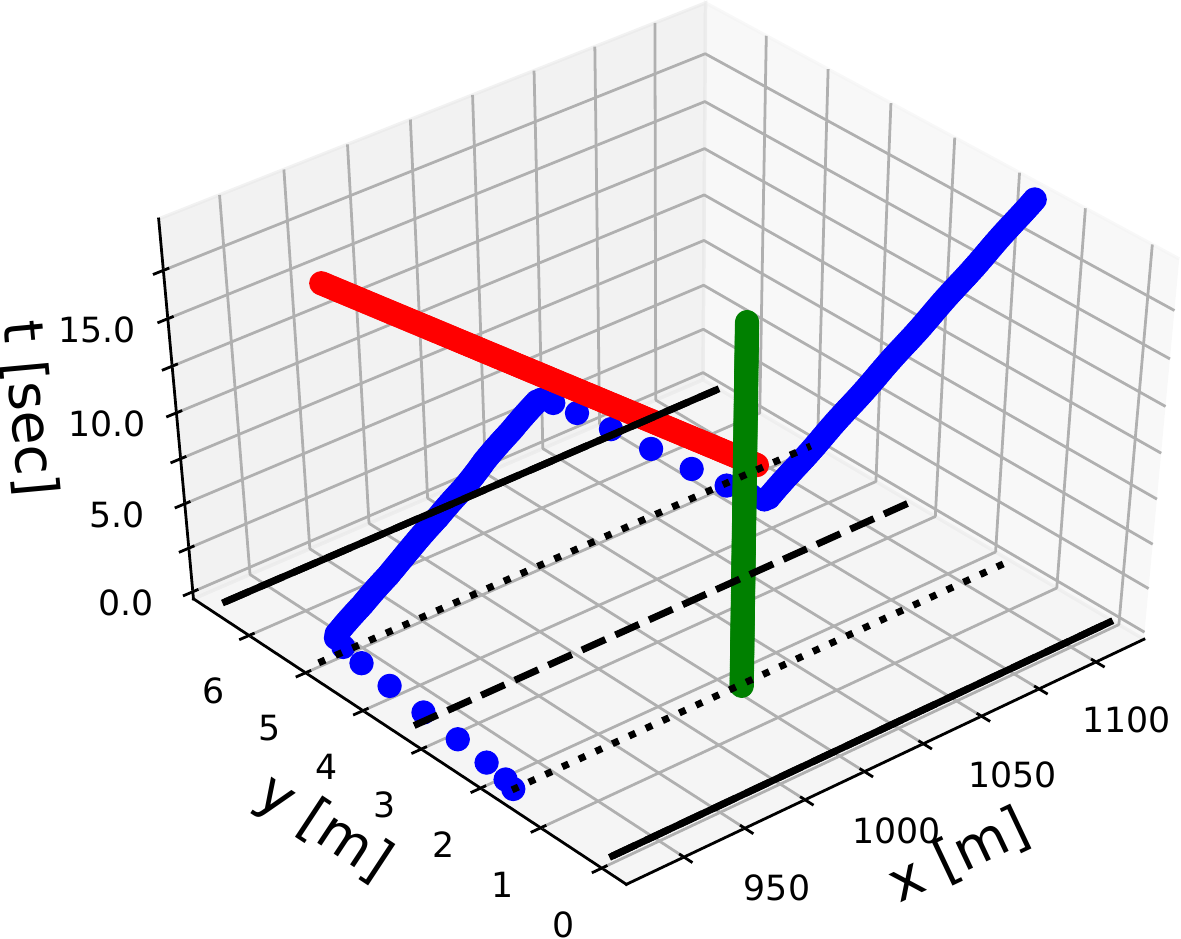}
    \caption{$v_{red} = \SI{9}{m/s}$}
  \end{subfigure}
  \begin{subfigure}{0.49\columnwidth}
  \includegraphics[width=\columnwidth]{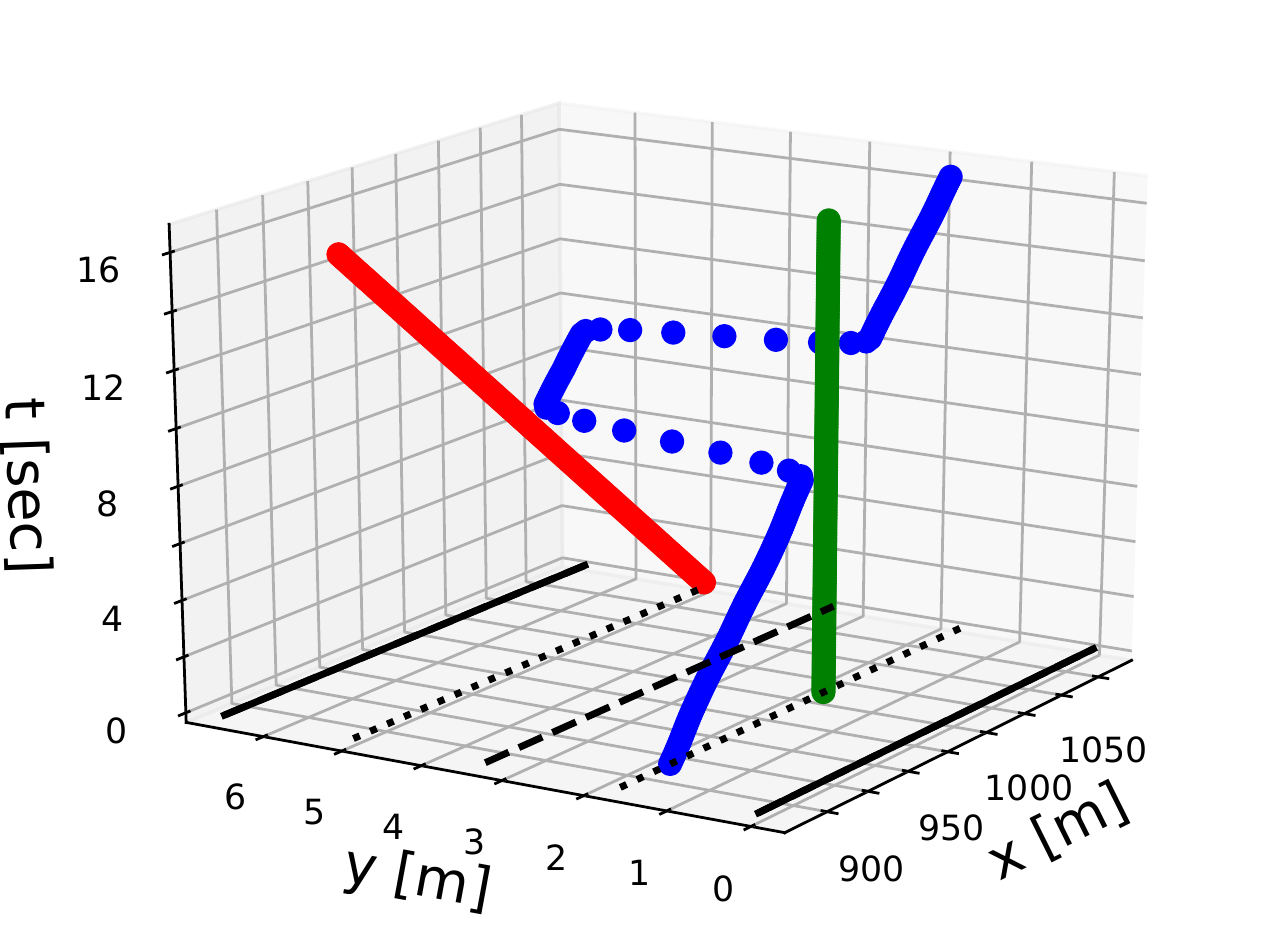}
  \caption{$v_{red} = \SI{13}{m/s}$}
  \end{subfigure}
  \begin{subfigure}{0.49\columnwidth}
  	\includegraphics[width=\columnwidth]{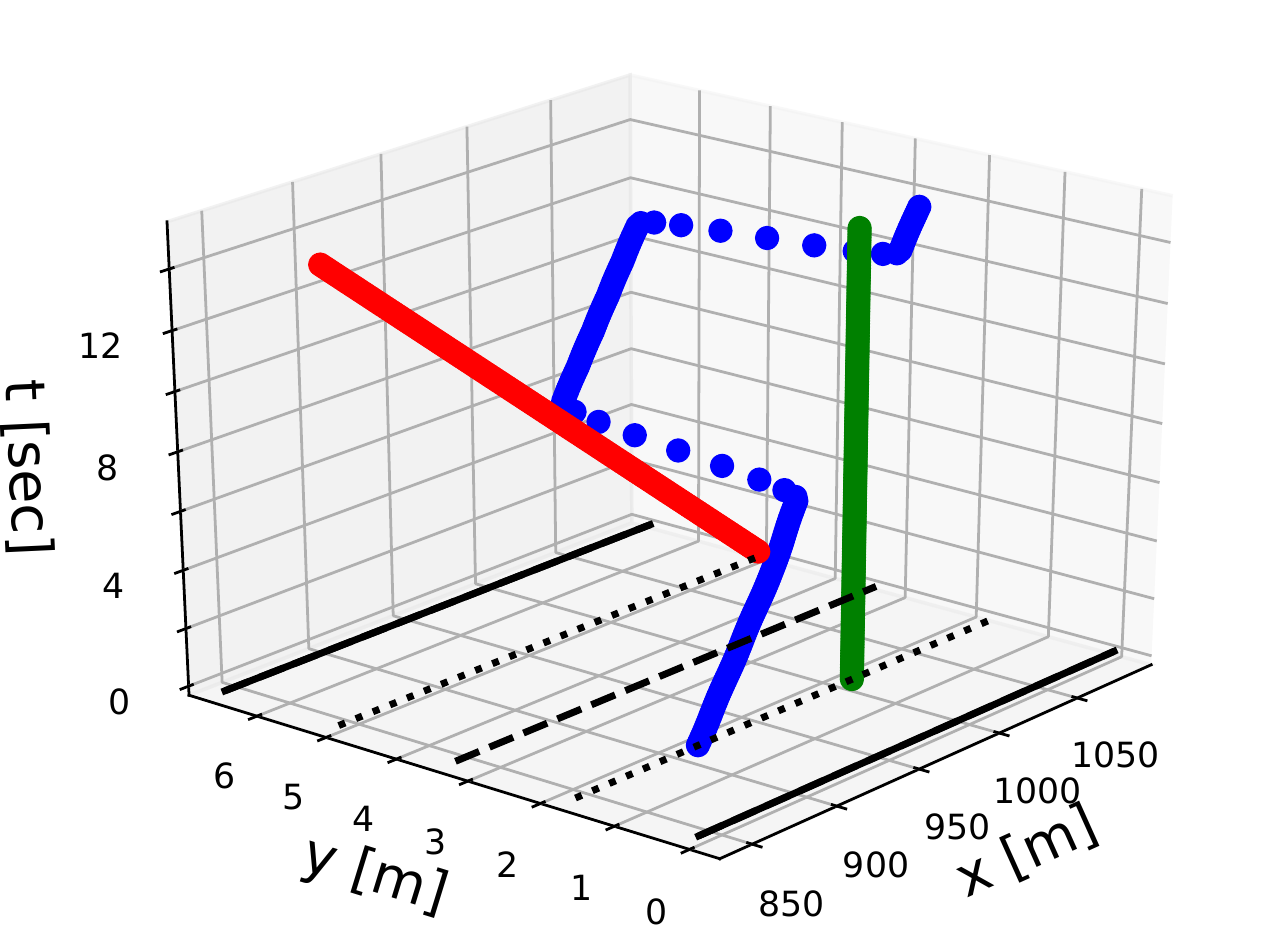}
    \caption{$v_{red} = \SI{17}{m/s}$}
    \end{subfigure}
  \caption{Different behavior for varying levels of cooperation}
  \label{Fig:BottleNeckEvaluation}
\end{figure}

Fig.~\ref{Fig:BottleNeckEvaluation} depicts the trajectories of the three vehicles for different velocities for vehicle 1.
When vehicle 1 drives at lower speeds, vehicle 0 chooses to drive faster to pass the bottleneck first.
When vehicle 1 drives at higher speeds, vehicle 0 changes its plan according to the current situation and lets the oncoming vehicle pass
before passing the obstacle.
It shows that our algorithm is able to generate robust solutions even in heterogeneous environments.
While the algorithm models others decisions as presented by ~\ref{sec:RewardFn},
it replans according to current situation in every steps in the \emph{polling} control mode.
Thus when another vehicle does not behave as assumed, \HMCTS\ will find a feasible plan accordingly.

\addtolength{\textheight}{-2cm}

\section{Conclusions}
In this paper, we proposed a decentralized planning method of MAs based on MCTS to generate cooperative maneuvers with longer time horizons. 
The hierarchical reinforcement learning framework \emph{MAXQ} is integrated into MCTS and then extended to the multi-agent system with the help of Decoupled-UCT.
By only specifying the initial and terminal conditions of MAs, the execution of MAs and the choice
over MAs are learned simultaneously.
The tests under several conflict scenarios show that our algorithm is able to handle a variety of conflict scenarios and
demonstrates potential over traditional MCTS.

Future work will focus on state abstraction, which will allow to share knowledge of macro-actions between different depths in the tree
as well as enable the recycling of the search tree, requiring even fewer iterations as possible future states have already been evaluated in previous plans.
Another aspect will be the integration of a learned prior distribution over actions as well as macro actions.

\section{Acknowledgements}
We wish to thank the German Research Foundation (DFG) for funding the project Cooperatively Interacting Automobiles (CoInCar)
within which the research leading to this contribution was conducted. 
The information as well as views presented in this publication are solely the ones expressed by the authors.




\listoftodos

\bibliographystyle{IEEEtran}
\bibliography{04_mendeley-export/library}

\end{document}

%% file: 03_graphics/main_mcts.pdf_tex
\begingroup%
  \makeatletter%
  \providecommand\color[2][]{%
    \errmessage{(Inkscape) Color is used for the text in Inkscape, but the package 'color.sty' is not loaded}%
    \renewcommand\color[2][]{}%
  }%
  \providecommand\transparent[1]{%
    \errmessage{(Inkscape) Transparency is used (non-zero) for the text in Inkscape, but the package 'transparent.sty' is not loaded}%
    \renewcommand\transparent[1]{}%
  }%
  \providecommand\rotatebox[2]{#2}%
  \ifx\svgwidth\undefined%
    \setlength{\unitlength}{1536bp}%
    \ifx\svgscale\undefined%
      \relax%
    \else%
      \setlength{\unitlength}{\unitlength * \real{\svgscale}}%
    \fi%
  \else%
    \setlength{\unitlength}{\svgwidth}%
  \fi%
  \global\let\svgwidth\undefined%
  \global\let\svgscale\undefined%
  \makeatother%
  \begin{picture}(1,0.44185772)%
    \put(0,0){\includegraphics[width=\unitlength,page=1]{main_mcts.pdf}}%
    \put(0.04068455,0.28532801){\color[rgb]{0,0,0}\makebox(0,0)[lb]{\smash{\fns Selection}}}%
    \put(0.28730513,0.28571686){\color[rgb]{0,0,0}\makebox(0,0)[lb]{\smash{\fns Expansion}}}%
    \put(0.53708459,0.28449501){\color[rgb]{0,0,0}\makebox(0,0)[lb]{\smash{\fns Simulation}}}%
    \put(0.76127374,0.28266223){\color[rgb]{0,0,0}\makebox(0,0)[lb]{\smash{\fns Backpropagation}}}%
  \end{picture}%
\endgroup%

%% file: 03_graphics/ActionGraph.pdf_tex
\begingroup%
  \makeatletter%
  \providecommand\color[2][]{%
    \errmessage{(Inkscape) Color is used for the text in Inkscape, but the package 'color.sty' is not loaded}%
    \renewcommand\color[2][]{}%
  }%
  \providecommand\transparent[1]{%
    \errmessage{(Inkscape) Transparency is used (non-zero) for the text in Inkscape, but the package 'transparent.sty' is not loaded}%
    \renewcommand\transparent[1]{}%
  }%
  \providecommand\rotatebox[2]{#2}%
  \ifx\svgwidth\undefined%
    \setlength{\unitlength}{604.03916016bp}%
    \ifx\svgscale\undefined%
      \relax%
    \else%
      \setlength{\unitlength}{\unitlength * \real{\svgscale}}%
    \fi%
  \else%
    \setlength{\unitlength}{\svgwidth}%
  \fi%
  \global\let\svgwidth\undefined%
  \global\let\svgscale\undefined%
  \makeatother%
  \begin{picture}(1,0.53310898)%
    \put(0,0){\includegraphics[width=\unitlength,page=1]{ActionGraph.pdf}}%
    \put(-0.00216226,0.30087457){\color[rgb]{0,0,0}\makebox(0,0)[lt]{\begin{minipage}{0.20298766\unitlength}\centering \fns to desired velocity\\ \end{minipage}}}%
    \put(0.22341778,0.28758045){\color[rgb]{0,0,0}\makebox(0,0)[lt]{\begin{minipage}{0.28605293\unitlength}\centering \fns make room \end{minipage}}}%
    \put(0.4905295,0.28938508){\color[rgb]{0,0,0}\makebox(0,0)[lt]{\begin{minipage}{0.28605293\unitlength}\centering \fns merge in \end{minipage}}}%
    \put(0.75847584,0.28757966){\color[rgb]{0,0,0}\makebox(0,0)[lt]{\begin{minipage}{0.28605293\unitlength}\centering \fns overtake\end{minipage}}}%
    \put(0.35735589,0.50091394){\color[rgb]{0,0,0}\makebox(0,0)[lt]{\begin{minipage}{0.28605293\unitlength}\centering \fns root\\ \end{minipage}}}%
    \put(0.22472529,0.21906749){\color[rgb]{0,0,0}\makebox(0,0)[lt]{\begin{minipage}{0.28605293\unitlength}\centering \fns $\pi_{mr}$\end{minipage}}}%
    \put(0.49183763,0.21906749){\color[rgb]{0,0,0}\makebox(0,0)[lt]{\begin{minipage}{0.28605293\unitlength}\centering \fns $\pi_{mi}$\end{minipage}}}%
    \put(0.75895011,0.21906749){\color[rgb]{0,0,0}\makebox(0,0)[lt]{\begin{minipage}{0.28605293\unitlength}\centering \fns $\pi_{ot}$\end{minipage}}}%
    \put(-0.04238726,0.21906749){\color[rgb]{0,0,0}\makebox(0,0)[lt]{\begin{minipage}{0.28605293\unitlength}\centering \fns $\pi_{dv}$\end{minipage}}}%
    \put(0.35828152,0.43240177){\color[rgb]{0,0,0}\makebox(0,0)[lt]{\begin{minipage}{0.28605293\unitlength}\centering \fns $\pi_{\mu}$\end{minipage}}}%
    \put(0.19113738,0.04256388){\color[rgb]{0,0,0}\makebox(0,0)[lt]{\begin{minipage}{0.16005923\unitlength}\centering \fns +\\ \end{minipage}}}%
    \put(0.42055693,0.04521272){\color[rgb]{0,0,0}\makebox(0,0)[lt]{\begin{minipage}{0.16005923\unitlength}\centering \fns 0\\ \end{minipage}}}%
    \put(0.53489415,0.04521272){\color[rgb]{0,0,0}\makebox(0,0)[lt]{\begin{minipage}{0.16005923\unitlength}\centering \fns L\\ \end{minipage}}}%
    \put(0.64998907,0.04521272){\color[rgb]{0,0,0}\makebox(0,0)[lt]{\begin{minipage}{0.16005923\unitlength}\centering \fns R\\ \end{minipage}}}%
    \put(0.30552175,0.03461738){\color[rgb]{0,0,0}\makebox(0,0)[lt]{\begin{minipage}{0.16005923\unitlength}\centering \fns -\\ \end{minipage}}}%
  \end{picture}%
\endgroup%

%% file: 03_graphics/PrimitiveActions.pdf_tex
\begingroup%
  \makeatletter%
  \providecommand\color[2][]{%
    \errmessage{(Inkscape) Color is used for the text in Inkscape, but the package 'color.sty' is not loaded}%
    \renewcommand\color[2][]{}%
  }%
  \providecommand\transparent[1]{%
    \errmessage{(Inkscape) Transparency is used (non-zero) for the text in Inkscape, but the package 'transparent.sty' is not loaded}%
    \renewcommand\transparent[1]{}%
  }%
  \providecommand\rotatebox[2]{#2}%
  \ifx\svgwidth\undefined%
    \setlength{\unitlength}{640.0003418bp}%
    \ifx\svgscale\undefined%
      \relax%
    \else%
      \setlength{\unitlength}{\unitlength * \real{\svgscale}}%
    \fi%
  \else%
    \setlength{\unitlength}{\svgwidth}%
  \fi%
  \global\let\svgwidth\undefined%
  \global\let\svgscale\undefined%
  \makeatother%
  \begin{picture}(1,0.30999369)%
    \put(0,0){\includegraphics[width=\unitlength,page=1]{PrimitiveActions.pdf}}%
    \put(0.73641439,0.29473341){\color[rgb]{0.60784314,0.80392157,0.8}\makebox(0,0)[lb]{\smash{\fns{\wb{lane change left}}}}}%
    \put(0.73641439,0.00390017){\color[rgb]{0.82745098,0.59215686,0.82745098}\makebox(0,0)[lb]{\smash{\fns{\wb{lane change right}}}}}%
    \put(0.81680233,0.11133382){\color[rgb]{0.6,0.80784314,0.59607843}\makebox(0,0)[lb]{\smash{\fns{\wb{accelerate}}}}}%
    \put(0.61611366,0.11133382){\color[rgb]{0.60392157,0.6,0.99607843}\makebox(0,0)[lb]{\smash{\fns{\wb{do nothing}}}}}%
    \put(0.39892974,0.11133382){\color[rgb]{0.99607843,0.61176471,0.61960784}\makebox(0,0)[lb]{\smash{\fns{\wb{decelerate}}}}}%
    \put(0.08390491,0.25884714){\color[rgb]{0,0,0}\makebox(0,0)[lb]{\smash{\fns{$\Delta y$}}}}%
    \put(0,0){\includegraphics[width=\unitlength,page=2]{PrimitiveActions.pdf}}%
    \put(0.24612811,0.01856507){\color[rgb]{0,0,0}\makebox(0,0)[lb]{\smash{\fns{$x$}}}}%
    \put(0.16750312,0.06155885){\color[rgb]{0,0,0}\makebox(0,0)[lb]{\smash{\fns{$y$}}}}%
  \end{picture}%
\endgroup%

%% file: 03_graphics/BoundedReturn.pdf_tex
\begingroup%
  \makeatletter%
  \providecommand\color[2][]{%
    \errmessage{(Inkscape) Color is used for the text in Inkscape, but the package 'color.sty' is not loaded}%
    \renewcommand\color[2][]{}%
  }%
  \providecommand\transparent[1]{%
    \errmessage{(Inkscape) Transparency is used (non-zero) for the text in Inkscape, but the package 'transparent.sty' is not loaded}%
    \renewcommand\transparent[1]{}%
  }%
  \providecommand\rotatebox[2]{#2}%
  \ifx\svgwidth\undefined%
    \setlength{\unitlength}{253.94901786bp}%
    \ifx\svgscale\undefined%
      \relax%
    \else%
      \setlength{\unitlength}{\unitlength * \real{\svgscale}}%
    \fi%
  \else%
    \setlength{\unitlength}{\svgwidth}%
  \fi%
  \global\let\svgwidth\undefined%
  \global\let\svgscale\undefined%
  \makeatother%
  \begin{picture}(1,0.52924137)%
    \put(0.74230806,0.44061599){\color[rgb]{0,0,0}\makebox(0,0)[lt]{\begin{minipage}{0.20685619\unitlength}\raggedright \fns overtake\\ \end{minipage}}}%
    \put(-0.31382803,0.43707389){\color[rgb]{0,0,0}\makebox(0,0)[lt]{\begin{minipage}{0.93361051\unitlength}\raggedleft \fns $g_{ot}^{\pi_\mu}=\gamma^0r_1+\gamma^1r_2+\gamma^2r_3+\gamma^3r_4$\end{minipage}}}%
    \put(0.74230806,0.36268153){\color[rgb]{0,0,0}\makebox(0,0)[lt]{\begin{minipage}{0.32590684\unitlength}\raggedright \fns lane change left\end{minipage}}}%
    \put(0.65125031,0.35661796){\color[rgb]{0,0,0}\makebox(0,0)[lt]{\begin{minipage}{0.16562471\unitlength}\raggedright \fns $r_{1}$\end{minipage}}}%
    \put(-0.31382665,0.35913526){\color[rgb]{0,0,0}\makebox(0,0)[lt]{\begin{minipage}{0.93361052\unitlength}\raggedleft \fns $g_{lane\,change\,left}^{\pi_{ot}}=\gamma^0r_1+\gamma^1r_2+\gamma^2r_3$\end{minipage}}}%
    \put(0.74230807,0.28607264){\color[rgb]{0,0,0}\makebox(0,0)[lt]{\begin{minipage}{0.23927527\unitlength}\raggedright \fns accelerate\end{minipage}}}%
    \put(0.65125031,0.28001017){\color[rgb]{0,0,0}\makebox(0,0)[lt]{\begin{minipage}{0.16562471\unitlength}\raggedright \fns $r_{2}$\end{minipage}}}%
    \put(-0.31382665,0.28252748){\color[rgb]{0,0,0}\makebox(0,0)[lt]{\begin{minipage}{0.93361052\unitlength}\raggedleft \fns $g_{accelerate}^{\pi_{ot}}=\gamma^0r_2+\gamma^1r_3$\end{minipage}}}%
    \put(0.74230807,0.21207413){\color[rgb]{0,0,0}\makebox(0,0)[lt]{\begin{minipage}{0.31869959\unitlength}\raggedright \fns lane change right\end{minipage}}}%
    \put(0.65125031,0.20601353){\color[rgb]{0,0,0}\makebox(0,0)[lt]{\begin{minipage}{0.16562471\unitlength}\raggedright \fns $r_{3}$\end{minipage}}}%
    \put(-0.31382661,0.20853084){\color[rgb]{0,0,0}\makebox(0,0)[lt]{\begin{minipage}{0.93361052\unitlength}\raggedleft \fns $g_{lane\,change\,right}^{\pi_{ot}}=r_3$\end{minipage}}}%
    \put(0.74230809,0.13283588){\color[rgb]{0,0,0}\makebox(0,0)[lt]{\begin{minipage}{0.36820336\unitlength}\raggedright \fns to desired velocity\end{minipage}}}%
    \put(-0.31382665,0.12929081){\color[rgb]{0,0,0}\makebox(0,0)[lt]{\begin{minipage}{0.93361052\unitlength}\raggedleft \fns $g_{dv}^{\pi_\mu}=r_4$\end{minipage}}}%
    \put(0.74230809,0.05229283){\color[rgb]{0,0,0}\makebox(0,0)[lt]{\begin{minipage}{0.23927527\unitlength}\raggedright \fns accelerate\end{minipage}}}%
    \put(0.65125031,0.04622795){\color[rgb]{0,0,0}\makebox(0,0)[lt]{\begin{minipage}{0.16562471\unitlength}\raggedright \fns $r_{4}$\end{minipage}}}%
    \put(-0.31382661,0.04874526){\color[rgb]{0,0,0}\makebox(0,0)[lt]{\begin{minipage}{0.93361052\unitlength}\raggedleft \fns $g_{accelerate}^{\pi_{dv}}=r_4$\end{minipage}}}%
    \put(0.74230809,0.52615657){\color[rgb]{0,0,0}\makebox(0,0)[lt]{\begin{minipage}{0.20685621\unitlength}\raggedright \fns root\\ \end{minipage}}}%
    \put(0,0){\includegraphics[width=\unitlength,page=1]{BoundedReturn.pdf}}%
  \end{picture}%
\endgroup%

%% file: root.bbl
\begin{thebibliography}{10}
\providecommand{\url}[1]{#1}
\csname url@rmstyle\endcsname
\providecommand{\newblock}{\relax}
\providecommand{\bibinfo}[2]{#2}
\providecommand\BIBentrySTDinterwordspacing{\spaceskip=0pt\relax}
\providecommand\BIBentryALTinterwordstretchfactor{4}
\providecommand\BIBentryALTinterwordspacing{\spaceskip=\fontdimen2\font plus
\BIBentryALTinterwordstretchfactor\fontdimen3\font minus
  \fontdimen4\font\relax}
\providecommand\BIBforeignlanguage[2]{{%
\expandafter\ifx\csname l@#1\endcsname\relax
\typeout{** WARNING: IEEEtran.bst: No hyphenation pattern has been}%
\typeout{** loaded for the language `#1'. Using the pattern for}%
\typeout{** the default language instead.}%
\else
\language=\csname l@#1\endcsname
\fi
#2}}

\bibitem{Bahram2016}
M.~Bahram, A.~Lawitzky, \emph{et~al.}, ``{A Game-Theoretic Approach to
  Replanning-Aware Interactive Scene Prediction and Planning},'' \emph{IEEE
  Transactions on Vehicular Technology}, 2016.

\bibitem{Elvik2014}
R.~Elvik, ``{A review of game-theoretic models of road user behaviour},''
  \emph{Accident Analysis and Prevention}, 2014.

\bibitem{Vodopivec2017}
T.~Vodopivec and B.~Ster, ``{On Monte Carlo Tree Search and Reinforcement
  Learning Spyridon Samothrakis},'' \emph{Journal of Artificial Intelligence
  Research}, 2017.

\bibitem{Silver2016}
D.~Silver, A.~Huang, \emph{et~al.}, ``{Mastering the game of Go with deep
  neural networks and tree search},'' \emph{Nature}, 2016.

\bibitem{Silver2017}
D.~Silver, J.~Schrittwieser, \emph{et~al.}, ``{Mastering the game of Go without
  human knowledge},'' \emph{Nature}, 2017.

\bibitem{Browne2012}
C.~B. Browne, E.~Powley, \emph{et~al.}, ``{A survey of Monte Carlo tree search
  methods},'' \emph{IEEE Transactions on Computational Intelligence and AI in
  Games}, 2012.

\bibitem{Kearns1999}
M.~Kearns, Y.~Mansour, and A.~Y. Ng, ``{A sparse sampling algorithm for
  near-optimal planning in large Markov decision processes},'' in \emph{IJCAI
  International Joint Conference on Artificial Intelligence}, 1999.

\bibitem{barto2003recent}
A.~G. Barto and S.~Mahadevan, ``Recent advances in hierarchical reinforcement
  learning,'' \emph{Discrete Event Dynamic Systems}, 2003.

\bibitem{Tak2014}
M.~J. Tak, M.~Lanctot, and M.~H. Winands, ``{Monte Carlo Tree Search variants
  for simultaneous move games},'' in \emph{IEEE Conference on Computatonal
  Intelligence and Games, CIG}, 2014.

\bibitem{Axelrod1981}
R.~Axelrod and W.~D. Hamilton, ``{The Evolution of Cooperation},''
  \emph{Science}, 1981.

\bibitem{During2014}
M.~D{\"{u}}ring and P.~Pascheka, ``Cooperative decentralized decision making
  for conflict resolution among autonomous agents,'' \emph{IEEE International
  Symposium on Innovations in Intelligent Systems and Applications}, 2014.

\bibitem{swaroop1996string}
D.~Swaroop and J.~K. Hedrick, ``String stability of interconnected systems,''
  \emph{IEEE Transactions on Automatic Control}, 1996.

\bibitem{stiller2007cooperative}
C.~Stiller, G.~Farber, and S.~Kammel, ``Cooperative cognitive automobiles,''
  \emph{IEEE, Intelligent Vehicles Symposium}, 2007.

\bibitem{Pascheka2015a}
P.~Pascheka and M.~During, ``{Advanced cooperative decentralized decision
  making using a cooperative reward system},'' in \emph{IEEE International
  Symposium on Innovations in Intelligent Systems and Applications}, 2015.

\bibitem{Takahashi1989}
A.~Takahashi, T.~Hongo, Y.~Ninomiya, and G.~Sugimoto, ``Local path planning and
  motion control for agv in positioning,'' in \emph{IEEE/RSJ International
  Conference on Intelligent Robots and Systems (IROS)}, Sept 1989.

\bibitem{Lenz2016}
D.~Lenz, T.~Kessler, \emph{et~al.}, ``{Tactical cooperative planning for
  autonomous highway driving using Monte-Carlo Tree Search},'' in \emph{IEEE
  Intelligent Vehicles Symposium}, 2016.

\bibitem{CowlingPeterI.2012}
{Cowling, Peter I.}, E.~J. Powley, and D.~Whitehouse, ``{Information set monte
  carlo tree search},'' \emph{IEEE Transactions on Computational Intelligence
  and AI in Games}, 2012.

\bibitem{Soemers2014}
D.~Soemers, ``{Tactical Planning Using MCTS in the Game of StarCraft},'' Ph.D.
  dissertation, Maastricht University, 2014.

\bibitem{Sutton1999a}
R.~S. Sutton, D.~Precup, \emph{et~al.}, ``{Between MDPs and semi-MDPs: A
  framework for temporal abstraction in reinforcement learning},''
  \emph{Artificial Intelligence}, 1999.

\bibitem{sutton1998intra}
R.~S. Sutton, D.~Precup, and S.~P. Singh, ``Intra-option learning about
  temporally abstract actions,'' in \emph{ICML}, 1998.

\bibitem{dietterich2000}
T.~G. Dietterich, ``Hierarchical reinforcement learning with the maxq value
  function decomposition,'' \emph{Journal of Artificial Intelligence Research},
  2000.

\bibitem{Powley2012}
E.~J. Powley, D.~Whitehouse, and P.~I. Cowling, ``{Monte Carlo Tree Search with
  macro-actions and heuristic route planning for the Physical Travelling
  Salesman Problem},'' in \emph{IEEE Conference on Computational Intelligence
  and Games}, 2012.

\bibitem{Perez2014}
D.~Perez, E.~J. Powley, D.~Whitehouse, P.~Rohlfshagen, S.~Samothrakis, P.~I.
  Cowling, and S.~M. Lucas, ``{Solving the physical traveling salesman problem:
  Tree search and macro actions},'' \emph{IEEE Transactions on Computational
  Intelligence and AI in Games}, 2014.

\bibitem{de2016monte}
M.~de~Waard, D.~M. Roijers, and S.~C. Bakkes, ``Monte carlo tree search with
  options for general video game playing,'' in \emph{IEEE Conference on
  Computational Intelligence and Games}, 2016.

\bibitem{Paxton2017a}
C.~Paxton, V.~Raman, \emph{et~al.}, ``{Combining Neural Networks and Tree
  Search for Task and Motion Planning in Challenging Environments},''
  \emph{IEEE/RSJ International Conference on Intelligent Robots and Systems},
  2017.

\bibitem{Vien2015}
M.~Toussaint, ``{Hierarchical Monte-Carlo Planning},'' in \emph{AAAI Conference
  on Artificial Intelligence}, 2015.

\bibitem{Bai2016a}
A.~Bai, S.~Srivastava, and S.~Russell, ``{Markovian state and action
  abstractions for MDPs via hierarchical MCTS},'' in \emph{IJCAI International
  Joint Conference on Artificial Intelligence}, 2016.

\bibitem{MahadevanSummary}
S.~Mahadevan, M.~Ghavamzadeh, \emph{et~al.}, ``{Hierarchical Approaches to
  Concurrency, Multiagency, and Partial Observability},'' \emph{Learning and
  Approximate Dynamic Programming: Scaling up to the Real World}, 2004.

\bibitem{Ghavamzadeh2006}
M.~Ghavamzadeh, S.~Mahadevan, \emph{et~al.}, ``{Hierarchical multi-agent
  reinforcement learning},'' \emph{Auton Agent Multi-Agent Sys}, 2006.

\bibitem{Liu2017}
M.~Liu, K.~Sivakumar, \emph{et~al.}, ``{Learning for Multi-robot Cooperation in
  Partially Observable Stochastic Environments with Macro-actions},'' 2017.

\bibitem{Amato2014a}
C.~Amato, G.~D. Konidaris, \emph{et~al.}, ``{Planning with Macro-Actions in
  Decentralized POMDPs},'' \emph{International conference on Autonomous Agents
  and Multi-Agent Systems}, 2014.

\bibitem{omidshafiei2015decentralized}
{Omidshafiei, Shayegan and Agha-Mohammadi, Ali-Akbar and others},
  ``{Decentralized control of partially observable markov decision processes
  using belief space macro-actions},'' in \emph{{International Conference on
  Robotics and Automation}}, 2015.

\bibitem{Ulbrich2015}
S.~Ulbrich, S.~Grossjohann, \emph{et~al.}, ``{Structuring Cooperative Behavior
  Planning Implementations for Automated Driving},'' in \emph{IEEE
  International Conference on Intelligent Transportation Systems}, 2015.

\bibitem{Lauer2000}
M.~Lauer and M.~Riedmiller, ``{An Algorithm for Distributed Reinforcement
  Learning in Cooperative Multi-Agent Systems},'' in \emph{International
  Conference on Machine Learning, ICML}, 2000.

\bibitem{ng1999policy}
A.~Y. Ng, D.~Harada, and S.~Russell, ``Policy invariance under reward
  transformations: Theory and application to reward shaping,'' in
  \emph{International Conference on Machine Learning, ICML}, 1999.

\bibitem{schaeffer2009comparing}
M.~Schaeffer, N.~Shafiei, \emph{et~al.}, ``Comparing uct versus cfr in
  simultaneous games,'' 2009.

\bibitem{NIPS2010_4031}
D.~Silver, J.~Veness, \emph{et~al.}, ``{Monte-Carlo Planning in Large
  POMDPs},'' in \emph{Advances in neural information processing systems, NIPS},
  2010.

\end{thebibliography}
